\documentclass[twoside,11pt]{article}

\usepackage{jmlr2e}

\usepackage{graphicx} 
\usepackage{enumitem, soul}
 \usepackage[dvipsnames*,svgnames,table,xcdraw]{xcolor}
\usepackage{adjustbox}
\newenvironment{bluebox}{%
    \noindent
    \adjustbox{innerenv={varwidth}[c]{0.9\linewidth},margin=\fboxsep+.25cm \fboxsep+.2cm,bgcolor=Bisque,frame,center}\bgroup
}{%
    \egroup
}

\usepackage{amsmath,amssymb,subcaption,multirow,amsthm}
\usepackage{adjustbox}
\usepackage{wrapfig,booktabs}
\usepackage{hyperref} 
\usepackage{mathtools}
\usepackage[]{algorithm2e}
\usepackage{hhline}
\usepackage{xspace}
\usepackage[tikz]{bclogo}
\newcommand{\muld}{DReaME}
\newcommand{\wrt}{w.r.t.}

\newtheorem{theorem}{Theorem}
\newtheorem{definition}{Definition}

\theoremstyle{remark}
\definecolor{ttblue}{RGB}{91,194,224}

\newcommand{\clB}{\mathcal{B}}
\newcommand{\clE}{\mathcal{E}}
\newcommand{\clEh}{{\hat \clE}}

\newcommand{\clH}{\mathcal{H}}

\newcommand{\z}{\mathrm{z}}

\newcommand{\kkb}{{\bar k}}
\makeatletter
\DeclareRobustCommand\onedot{\futurelet\@let@token\@onedot}
\def\@onedot{\ifx\@let@token.\else.\null\fi\xspace}
\def\eg{\emph{e.g}\onedot} 
\def\ie{\emph{i.e}\onedot}

\def\wrt{w.r.t\onedot} 
\def\etal{\emph{et al}\onedot}
\makeatother
\def\x{{\mathbf x}}

\def\gro{{GroupDRO}\xspace}

\def\regmethod{{GroupDRO++}\xspace}

\firstpageno{1}

\begin{document}

\title{Automated Domain Discovery from Multiple Sources to Improve Zero-Shot Generalization}

\author{\name Kowshik Thopalli
\email kthopall@asu.edu \\
      \addr Geometric Media Lab, \\Department of ECEE, \\
      Arizona State University,\\
      Tempe, AZ, USA\\
      \AND
      \name Sameeksha Katoch \email skatoch1@asu.edu\\
      \addr Qualcomm\\
      San Diego, CA, USA \\
      \AND
      \name Pavan Turaga \email pturaga@asu.edu\\
      \addr Geometric Media Lab, \\Department of ECEE, \\
      Arizona State University,\\
      Tempe, AZ, USA\\
      \AND
      \name Jayaraman J.\ Thiagarajan \email jayaramanthi1@llnl.gov \\
      \addr  Lawrence Livermore National Laboratory\\
      Livermore, CA, USA\\
      }



            


\maketitle

\begin{abstract}
Domain generalization (DG) methods aim to develop models that generalize to settings where the test distribution is different from the training data. In this paper, we focus on the challenging problem of multi-source zero shot DG (MDG), where labeled training data from multiple source domains is available but with no access to data from the target domain. A wide range of solutions have been proposed for this problem, including the state-of-the-art multi-domain ensembling approaches. Despite these advances, the na\"ive ERM solution of pooling all source data together and training a single classifier is surprisingly effective on standard benchmarks. In this paper, we hypothesize that, it is important to elucidate the link between pre-specified domain labels and MDG performance, in order to explain this behavior. More specifically, we consider two popular classes of MDG algorithms -- distributional robust optimization (DRO) and multi-domain ensembles, in order to demonstrate how inferring custom domain groups can lead to consistent improvements over the original domain labels that come with the dataset. To this end, we propose (i) \regmethod, which incorporates an explicit clustering step to identify custom domains in an existing DRO technique; and (ii) \muld~, which produces effective multi-domain ensembles through implicit domain re-labeling with a novel meta-optimization algorithm. Using empirical studies on multiple standard benchmarks, we show that our variants consistently outperform ERM by significant margins ($1.5\% - 9\%$), and produce state-of-the-art MDG performance.
Our code can be found at \url{https://github.com/kowshikthopalli/DREAME}
\end{abstract}

\begin{keywords}
Multi-domain generalization, distribution shifts, ERM, domain re-labeling, ensembles, distributional robust optimization, meta learning
\end{keywords}

\section{Introduction}
\label{sec:intro}
 Supervised machine learning models commonly rely on the assumption that the training and testing data are independent and identically distributed (\textit{i.i.d.}). As a result, such models can fail drastically when tested on data that is not \textit{i.i.d.} ~\citep{torralba2011unbiased}.
Such drops in performance indicate poor generalization capabilities of the models. Addressing this fundamental challenge has become an important topic of research~\citep{cycada,DANN,deng2018image} over the last few years. Several classes of solutions, ranging from unsupervised adaptation approaches~\citep{wang2018deep}, new formulations for domain-invariant learning~\citep{ARM, arjovsky2019invariant}, data augmentation techniques~\citep{randconv} and novel regularization strategies~\citep{mixstyle}, have been proposed to improve generalization under covariate shifts. Furthermore, given the inherent limitation of using data from a single source domain to generalize under real-world shifts, methods that leverage multiple domains have also emerged~\citep{blanchardDG_thm}. Commonly referred to as zero-shot, multi-domain generalization (ZS-MDG), this formulation assumes that labeled data from multiple source domains is available but with no access to the target domain.

The simplest solution to this problem is the vanilla empirical risk minimization (ERM)~\citep{vapnik1999overview}, which minimizes an average loss computed on data pooled together from all available source domains. The inability of this approach to exploit statistical discrepancies between domains has motivated the design of multi-domain learning techniques~\citep{survey_dg}. However, recently Gulrajani \textit{et al.}~\citep{gulrajani2020search} reported that a powerful feature extractor coupled with effective model selection can make ERMs highly competitive on standard benchmarks. Since then there is renewed interest in better understanding and improving the performance of vanilla ERM. In this context, approaches that enforce ERM-based models to be consistent under appropriate data augmentations have become popular~\citep{randconv, mbdg}. Despite their effectiveness, choosing the most appropriate augmentation for a given dataset is challenging, and in practice, even advanced strategies, e.g., random convolutions, can provide varying degrees of performance gains across datasets.

In this paper, we take a different perspective for improving ZS-MDG by exploit the important link between domain-aware MDG solutions and the domain groups in a dataset. To this end, we first consider a recent distributional robust optimization algorithm (GroupDRO~\citep{GroupDRO}), wherein we develop a new variant \regmethod that jointly infers optimal ``domain'' groups (through a clustering step) and performs multi-source model training with GroupDRO. Our analysis shows that, \regmethod outperforms both the vanilla ERM and the standard GroupDRO implementation on all benchmarks.



Next, we consider another popular class of MDG methods, multi-domain ensembles. Existing ensembling methods~\citep{DMG,DSON} typically combine multiple models (that can optionally share parameters) that are trained for each of the observed domains. However, it is non-trivial to use an approach like \regmethod in this case, since \regmethod requires integration of an auxiliary clustering step (e.g., K-means) and needs significant modification to the training process for joint optimization. Hence, We introduce \muld~(\ul{D}omain \ul{Re}-l\ul{a}beling for \ul{M}ulti-Domain \ul{E}nsembles), that advances existing multi-domain ensembling techniques by implicitly reorganizing the data samples into custom domain groupings during ensemble construction, through a novel meta optimization algorithm. Using extensive empirical studies on a large suite of multi-domain benchmarks, we argue that, one can learn invariances that are the most beneficial for out-of-domain generalization through a more meaningful re-grouping of data samples. Our results show that, the proposed approaches lead to significant performance gains over the ERM baseline, and more importantly, produce state-of-the-art performance on challenging datasets.


\subsection{Contributions and Findings}

We make the following contributions and findings in regard to ZS-MDG: (i) We hypothesize that domain groups play a critical role in determining the performance of MDG algorithms and validate the hypothesis with two popular MDG approaches; (ii) We introduce GroupDRO++ to infer ``optimal'' domain labels in a distributional robust optimization setting; (iii) Next, we propose \muld, a new multi-domain ensembling method with implicit domain re-labeling. \muld~uses a gradient-matching based re-labeling strategy that is found to be empirically superior to other design choices; (iv) We investigate two model selection protocols for \muld~using only the source domain validation data; and (v) We perform extensive empirical studies with DomainBed~\citep{gulrajani2020search} and a large suite of benchmarks (OfficeHome, Camelyon$17$-WILDS, PACS and Terra-Incognita), to demonstrate the effectiveness of our proposed approaches.

\subsection{Paper Organization}
The rest of this paper is organized as follows. Section 2 discusses the background on multi-domain generalization and known theoretical bounds on generalization error in comparison to standard ERM. Next, in Section 3, we present our hypothesis on the need for domain re-labeling. Sections 4 and 5 present our proposed algorithms that integrate automated domain discovery into two popular MDG algorithms, \gro and multi-domain ensembles, respectively. Section 6 presents the empirical findings and a discussion on related work is provided in Section 7. Finally, Section 8 provides concluding remarks.

\section{Background}
\label{sec:background}
\paragraph{Problem Setup}Given access to $K$ labeled source domains $\{\mathcal{D}_{1}, \ldots, \mathcal{D}_{K}\}$ where $\mathcal{D}_{k} = \{(\x_{i}^k, y_{i}^k)\}_{i = 1}^{N_{k}} \sim P^{(k)}_{XY}$ is the $k^{th}$ domain comprising $N_{k}$ image-label pairs, the goal is to generalize to any novel test domain $\mathcal{D}^{\dagger}$, without requiring labeled or unlabeled examples.
We consider the homogeneous MDG setting \textit{i.e.}, observed and unobserved domains share the same label space.

\paragraph{Vanilla ERM} Here, the goal is to find a function $f:\x \rightarrow y$ that maps samples $\x$ to labels $y$ by minimizing the empirical risk on the pooled data:
\begin{equation}
	\vspace{-2pt}
	\frac{1}{|\mathcal{D}|} \sum_{(\x_i,y_i) \in \mathcal{D}} \ell\left(f\left(\x_{i}\right), y_{i}\right); \quad \mathcal{D} \coloneqq \bigcup_{k=1}^K \mathcal{D}_k,
	\label{eq:erm}
\end{equation}
where, $\ell$ is the loss function that measures predictive error, \textit{e.g.}, cross-entropy, and $|\mathcal{D}|$ denotes the size of the pooled dataset.

Since this simple baseline method does not leverage the inherent discrepancies between the sources domains, one might expect this to be ineffective in practice. Surprisingly, this na\"ive solution has been found to produce competitive performance on standard domain generalization benchmarks~\citep{gulrajani2020search}. Consequently, there is a critical need to understand why MDG methods that leverage the knowledge about the domain groups do not provide non-trivial gains over ERM.

\paragraph{Improving ERM by Maximizing Worst-case Performance}A major drawback of the formulation in~\eqref{eq:erm} is that it treats all samples from all domains/groups equally and thus decreases the loss in an average sense. An insight from distributional robust optimization is that decreasing the error on worst-group can lead to better generalization. Hence, an important class of MDG algorithms attempts to decrease a weighted mean of the group-level (pre-specified in the dataset) losses with an adaptive domain-specific weighting, such that large weights are assigned to groups with higher error. Formally, let $g_i = k \in \{1, \cdots, K\}$ denote the group to which a sample $(\x_i, y_i)$ belongs to, where $K$ is the total number of groups, and $q_k$ is the weight assigned to a particular group. The risk now becomes
\begin{equation}
	\frac{1}{|\mathcal{D}|} \sum_{(x_i,y_i,g_i) \in \mathcal{D}} q_{g_i} \ell\bigg(f(x_{i}), y_{i}\bigg); \quad \mathcal{D} \coloneqq \bigcup_{k=1}^K \mathcal{D}_k.
	\label{eq:groupDRO}
\end{equation}For example, GroupDRO~\citep{GroupDRO} uses an update rule for $q_g$ across iterations, such that a group with larger error is assigned a higher weight.

\paragraph{Error Bounds for MDG}Before introducing our hypothesis on the crucial role that domain labels play, we briefly discuss existing theoretical bounds for MDG in a binary classification setting~\citep{blanchardDG_thm}. Let us begin by defining a hyper-distribution $\bar{P}$ on $(\x, y)$ from which all source domains, $P_{XY}^{(k)}, k = 1, \cdots, K$, and the target domain, $P_{XY}^{\dagger}$, are drawn from. In order to train a classifier $f$ that generalizes to any possible target domain, one can leverage the knowledge of domain groupings during training. In such a case, the average risk can be estimated as follows:
\begin{equation}
    \mathcal{E}(f) \coloneqq \mathbb{E}_{P_{XY} \sim \bar{P}} \text{ } \mathbb{E}_{(\x, y) \sim P_{XY}} \ell\bigg(f(P_X, \x), y\bigg),
\end{equation}where $P_X$ is the marginal distribution for a domain drawn from $\bar{P}$. In practice, we use finite approximations for these expectations:
\begin{equation}
    \hat{\mathcal{E}}(f) \coloneqq \frac{1}{K} \sum_{k=1}^K \frac{1}{N_k} \sum_{i=1}^{N_k} \ell\bigg(f(\mathcal{X}_k, \x_i^k), y_i^k\bigg), \mathcal{X}_k = \{\x_i^k | (\x_i^k, y_i^k) \in \mathcal{D}_k\}.
\end{equation}The discrepancy between $\mathcal{E}(f)$ and the estimate $\hat{\mathcal{E}}(f)$ can be measured using an appropriate divergence in some hypothesis space for $f$. For example, Blanchard \textit{et al.}~\citep{blanchardDG_thm} considered the space of $f$ to be a reproducing kernel Hilbert space (RKHS), where the inducing kernel is of the form $\mathrm{k}((P_X^{(m)}, \x_i^m), (P_X^{(n)}, \x_j^n))$. The following theorem provides an upper bound on the discrepancy $|\hat{\mathcal{E}}(f) - {\mathcal{E}}(f)|$:

\begin{theorem}[Average risk estimation error bound for binary classification~\citep{blanchardDG_thm, survey_dg}]
    Assume that the loss function $\ell$ is $L_\ell$-Lipschitz in its first argument and is bounded by $B_\ell$.
    Assume also that the kernels $\mathrm{k}_X, \mathrm{k}^{\prime}_X$ and $\kappa$ are bounded by $B_{\mathrm{k}}^2, B_{\mathrm{k}^{\prime}}^2 \ge 1$ and $B_\kappa^2$, respectively, and the canonical feature map $\Phi_\kappa: v \in \clH_{\mathrm{k}^{\prime}_X} \mapsto \kappa(v, \cdot) \in \clH_\kappa$ of $\kappa$ is $L_\kappa$-H\"older of order $\alpha \in (0,1]$ on the closed ball $\clB_{\clH_{\mathrm{k}^{\prime}_X}} (B_{\mathrm{k}^{\prime}})$ \footnote{
        This means that for any $u,v \in \clB_{\clH_{\mathrm{k}^{\prime}_X}} (\clB_{\clH_{\mathrm{k}^{\prime}}})$, it holds that $\Vert \Phi_\kappa(u) - \Phi_\kappa(v) \Vert \le L_\kappa \Vert u - v \Vert^\alpha$, where the norms are of the respective RKHSs.}.
    Then for any $r > 0$ and $\delta \in (0,1)$, with probability at least $1-\delta$, it holds that:
    \begin{align}
        \sup_{f \in \clB_{\clH_\kkb} (r)} \left| \clEh(f) - \clE(f) \right|
        \le{} & C \bigg( B_\ell \sqrt{- K^{-1} \log \delta} \\
        + r B_{\mathrm{k}} L_\ell \Big( B_{\mathrm{k}^{\prime}} L_\kappa \big( & n^{-1} \log (K/\delta) \big)^{\alpha/2} + B_\kappa / \sqrt{K} \Big) \bigg),
    \end{align}
    where $C$ is a constant.
\end{theorem}For simplicity, assuming that $\forall N_k = N$, this upper bound becomes larger when $(K, N)$ is replaced with $(1, KN)$ thus indicating that using domain groupings leads to lower error than pooling all of them as done by ERM~\citep{survey_dg}. Under the light of this result, the success of ERM over methods that use domain labels is very surprising and warrants attention.

\section{Need for Domain Re-labeling}
\label{sec:motivation}
An overarching assumption made by most existing MDG methods is that exploiting the domain groups (e.g., photos, art, cartoon, sketch in PACS~\cite{PACS} dataset) will lead to improved generalization. However, in practice, we find that these methods do not consistently provide non-trivial gains over ERM, particularly when fair model selection strategies are adopted~\cite{gulrajani2020search}. Hence, in this paper, we hypothesize that the domain labels play a crucial role in determining the trade-off between transferability (\ie domain invariance features) and domain-level performance (\ie worst group error). Formally, we state our hypothesis as follows:
\vspace{0.1in}

\begin{bluebox}
   \textbf{Hypothesis:} \textit{Existing MDG approaches that exclusively optimize for minimizing the discrepancy between pre-specified domain groups (i.e., transferability) present the risk of compromising the performance in one or more of the source domains. Hence, we hypothesize that by constructing custom domain groups one can effectively trade-off transferability and domain-specific performance.}
\end{bluebox}
\vspace{0.1in}

Consequently, we propose to improve existing domain-label aware learning techniques (\gro and multi-domain ensembles in this paper) by including a custom domain discovery step into the training process. To this end, we consider two popular MDG algorithms -- GroupDRO~\cite{GroupDRO}, a distributional robust optimization technique, and multi-domain ensembles. At a high-level, we ignore the domain-labels that come with the dataset  and attempt to re-categorize the data into custom domain groups during training. We describe in detail our proposed methods, \regmethod and \muld~in Sections 4 and 5 respectively.

\section{Improving \gro with Custom Domain Discovery}
\label{sec:gdro++}

\RestyleAlgo{boxruled}
\begin{algorithm}[t]
	\KwIn{Set of training domains $\mathcal{D}\coloneqq\{\mathcal{D}_1\dots\mathcal{D}_K\}$, 
	hyper-parameters $\lambda, \gamma$, $\eta_q$, $\alpha$\
	}
	\KwOut{Trained model $f({\theta})= h \circ c$ with feature extractor $h$ and classifier $c$}
	\textbf{Initialization}: Model parameters $\theta^{(0)}$, group label for each sample $G \coloneqq \{g_i\}$ and weights $\{q_k\}$   \;
	\For{iter \textbf{in} $n_{iter}$}
	{
	\For{t \textbf{in} $1\cdots T$}
	{
	        //run \gro \\
		       \For {$k$ in $1 \cdots K$}
		      {
		       //update group weight
		      $ \bar{q}_{k} \leftarrow q_k \exp(\eta_{q} \ell(\theta^{(t-1)} ;(\x, y)))$ 
		      } 
		   //Renormalize q \\
            $q_k \leftarrow \frac{\bar{q}_k}{\sum_{l} \bar{q}_l}$\; 
            $ \displaystyle\mathcal{L}=\sum_{k}q_{k} \sum_{\{(\x_i, y_i)|g_i = k\}}\ell\bigg(f(\x_i), y_i \bigg)$\; 
            $\displaystyle\mathcal{R}=\sum_i q^{\gamma}_{g_i} \ell\bigg(f(\x_i), y_i\bigg)$\;
            //update $\theta$ \\
           $\theta^{(t)} \leftarrow \theta^{(t-1)}-\alpha \nabla_{\theta} (\mathcal{L}+\lambda\mathcal{R})$
    
    }        

  //Cluster the samples//\\
		      $\mathcal{Z}\coloneqq\cup_i h(\x_i)$\;
              $G \leftarrow$ $K$-means$(\mathcal{Z}) $ 
		 
}
	\caption{GroupDRO++}\label{algo:algo1}
\end{algorithm}

While one can design a variety of approaches to group samples, we adopt a deep clustering-based solution~\citep{deepcluster} that iteratively clusters latent representations via $K$-means and uses the cluster-labels as domain-labels to perform GroupDRO training. We begin by first extracting the latent representations of data from all source domains via a pre-trained feature extractor $h$. We represent the features using the set $\mathcal{Z}\colon\{\z_1,\z_2....\z_{N}\}$  with $\z_i = h(\x_i)$, where $N$ denotes total number of samples pooled from all source domains. Subsequently, $\mathcal{Z}$ is clustered using $K$-Means to form $M$ groups (can be equal to or different than $K$) and each sample can now be represented as a tuple $(\x_i,y_i,g_i)$, where $g_i$ denotes the cluster label. The feature extractor $h$ and the classifier $c$ are trained with a \gro style optimization \ie, with adaptive weights for each group for a pre-specified number of iterations $T$, following which, we re-compute the latent representations for data using the updated model and perform clustering to refine the group labels. 

\begin{figure}[t]
    \centering
    \includegraphics[width=0.95\columnwidth]{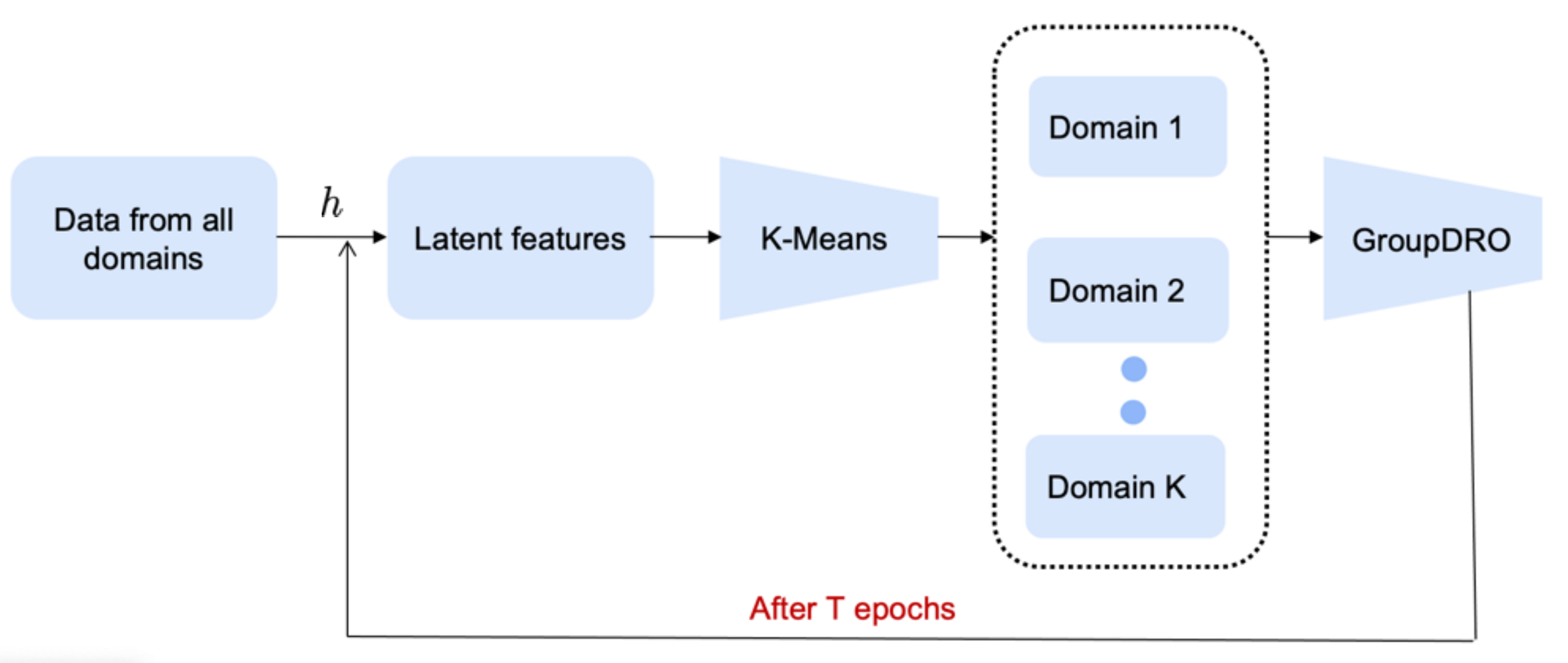}
    \caption{\textbf{\regmethod Overview}. An illustration of our approach for  generating custom domain groups to improve the \gro technique.}
    \label{fig:groupdro++}
\end{figure}

Since the clustering algorithm is disconnected from the model training except when utilizing the updated features, it is important to regularize the training so that meaningful domain groups can be created. Note, ERM aggregates losses at the sample-level, while \gro operates at the group level, and there is a need to enable a finer control. To illustrate this intuitively, consider these two non-desirable cases: (i) An individual sample can have a high loss while its group has been assigned a smaller weight. 
In this case, the update to the model $f = h \circ c$ via \gro will not have the desired effect as the weight is low; (ii) On the other hand, a sample having a low loss value when its group has a larger weight. Due to this wrong group association, that sample would still contribute to the SGD update. To address these issues, we introduce a regularization term that balances both the group-level weighting and sample-level weighting: 
$$\mathcal{R}=\ell(f(\z_i), y_i) q^{\gamma}_{g_i}, \quad 0<\gamma<1.$$Here $\ell(f(\z_i), y_i)$ is the sample level mis-classification error, $q_{g_i}$ is the weight assigned to the group that the sample belongs to and $\gamma$ controls the sharpness of the regularizer. Thus, the final objective is given by $\ell(f(\z_i), y_i) +\lambda \mathcal{R}$, where $\lambda$ is the weight associated with the regularization term when attempting to maximize worst-case performance. Our algorithm is summarized in Algorithm~\ref{algo:algo1}.

\begin{figure*}[t!]
    \centering
    \includegraphics[width=0.75\textwidth]{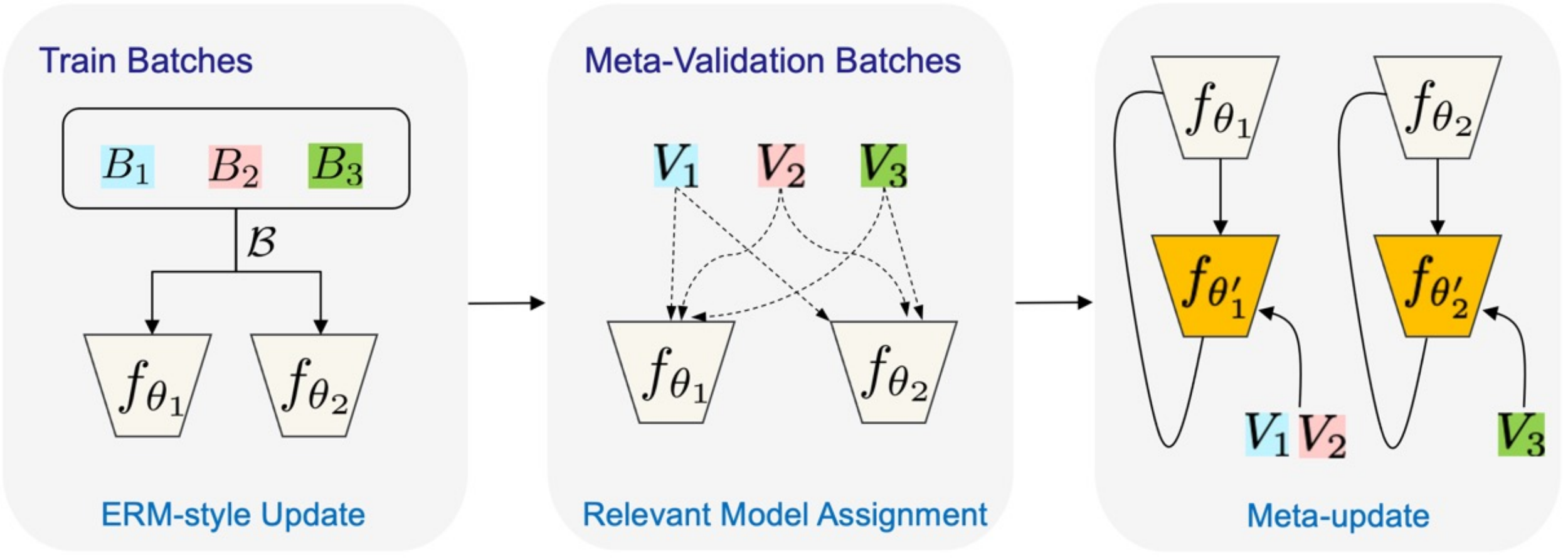}
\caption{\textbf{\muld~overview}. In the meta-train stage, we compute ERM-style gradients for all models $M_i$ using the data batch $\mathcal{B} \coloneqq \{\mathcal{B}_1,\mathcal{B}_2,\mathcal{B}_3\}$~pooled from all source domains. Next, we use a novel model relevance score computation to assign each meta-validation batch to the most relevant model in the ensemble. Finally, we use a gradient-through-gradient update to learn each of the models $f_{\theta_i}$.
}
    \label{fig:intro_fig}
\end{figure*}

\section{Domain Re-Labeling for Multi-Domain Ensembles}
\label{sec:Dreame}
In this section, we describe our approach for incorporating domain discovery in multi-domain ensembles. 

Formally, we represent a multi-domain ensemble using a set of models $\left\{f_{\theta_{m}}\right\}_{m=1}^{M}$, where $M$ is the ensemble size (typically set to the number of source domains $K$) and all models are initialized randomly. Upon training, the predictions for samples from a novel target domain $\mathcal{D^\dagger}$ can be obtained as an unweighted average of predictions, \ie, $\frac{1}{M} \sum_{m=1}^{M} f_{\theta_m}(\x_i), \forall \x_i \in \mathcal{D^\dagger}$. Labeled data from each observed domain $\mathcal{D}_k$ is divided into three disjoint sets - train $\mathcal{D}_k^{t}$, meta-validation $\mathcal{D}_k^{v}$ and held-out validation $\mathcal{D}_k^{hv}$. Note, in contrast to existing approaches, $M$ can be different from $K$ in \muld.

Our approach is comprised of two stages, both operating at the mini-batch level - (i) \textbf{meta-train stage}: Obtain ERM-style gradients for each constituent member of the ensemble using the collection $\bigcup_k \mathcal{D}_k^{t}$ from all $K$ source domains; (ii) \textbf{meta-test stage}: Utilize the proposed \textit{model relevance score} (MRS) to determine the most appropriate model $f_{\theta_m}$ from the ensemble, to apply for each of the meta-validation sets $\mathcal{D}_k^{v}$ and subsequently $f_{\theta_m}$ is updated only using meta-gradients from the subset of $\mathcal{D}_k^{v}$'s assigned to this model. This step enables the implicit re-organization of the validation mini-batches into different domain groups. In addition, we also explore the use of synthetic augmentations to create additional meta-validation batches for exposing the models to more diverse variations of data from the observed domains. Finally, the held-out validation sets $\{\mathcal{D}_k^{hv}\}$ are used for model selection (following standard practice). Algorithm~\ref{algo:algo2} lists the steps involved in our algorithm.

\subsection{Meta-train Stage: } 
\label{sec:meta-train}
In every iteration, $K$ mini-batches $\{\mathcal{B}_k^t\}$ are randomly sampled from the $K$ training sets $\{\mathcal{D}_{k}^t\}$, which are then pooled to form the data batch $\mathcal{B} \coloneqq \bigcup_k \mathcal{B}_k^t \subset \mathcal{D}_{k}^t$ and passed as input to all $M$ models (initialized randomly). The empirical risk
\begin{equation}  \mathcal{L}_{\theta_m}=\frac{1}{|\mathcal{B}|} \sum_{(\x_i,y_i) \in \mathcal{B}} \ell(f_{\theta_m}(\x_{i}), y_{i}), \forall m \in (1,\cdots,M),\label{eq:meta-train-erm}
\end{equation} and the corresponding gradients for each of the models $\nabla_{\theta_m}(\mathcal{L}_{\theta_m})$ are computed independently \wrt~$\mathcal{B}$. Akin to any MAML~\citep{finn2017model} style algorithm, \muld~takes one gradient step for each of the models $f_{\theta_m}$ to obtain $f_{\theta^{\prime}_m}$, \ie,
\begin{equation}
\theta^{\prime}_m = \theta_m - \alpha  \nabla_{\theta_m} \mathcal{L}_{\theta_m}(\theta_m), \forall m \in (1,\cdots,M)
\label{eqn: inner_loop_update}
\end{equation}with a pre-specified learning rate $\alpha$. 
 
\RestyleAlgo{boxruled}
\begin{algorithm}[t]
	\KwIn{Set of training domains $\mathcal{D}\coloneqq\{\mathcal{D}_1\dots\mathcal{D}_K\}$}
	\KwOut{ Ensemble $\{f_{\theta_{1}}, \dots ,f_{\theta_{M}}\}$}
	\textbf{Initialization}: Parameters $\{\theta_{1}, \dots ,\theta_{M}\}$,
	meta-train sets $\{\mathcal{D}^{t}_1\dots\mathcal{D}^{t}_K\}$, meta-validation sets $\{\mathcal{D}^{v}_1\dots\mathcal{D}^{v}_K\}$, 
	hyper-parameters $\alpha,\lambda, \eta$ \;

	\For{iter \textbf{in} $n_{iter}$}{
	        //meta-train // \\
		       \For {$f_{\theta_m}$ in $f_{\theta_1} \cdots f_{\theta_M}$}{
		              $\mathcal{B} = [\mathcal{B}_{1}, \cdots \mathcal{B}_{K}]$ // pool the minibatches from $\{\mathcal{D}_{k}^t\}$  //
		              
		              Compute empirical risk ~$\mathcal{L}_{\theta_m}$ \wrt~$\mathcal{B}$, using eq. ~\eqref{eq:meta-train-erm}\;
		              //inner gradient update// \\
		              Update $\theta^{\prime}_m = \theta_m - \alpha  \nabla_{\theta_m} \mathcal{L}_{\theta_m}(\theta_m)$  
		             } 
		             
             //model relevance score//\\
		      \For{$k$ in $1 \cdots K $ }{
		           \For{$f_{\theta_m}$ in $f_{\theta_1} \cdots f_{\theta_M}$}{
		                sample a mini-batch $\mathcal{V}_k$ from $\mathcal{D}_{k}^v$\;
		                compute $\beta_{km}$ using eq. ~\eqref{eqn:BMRS}\;
		                }
		          }   
		          
		      //meta-update//\\
		      
		      \For{$f_{\theta_m}$ in $f_{\theta_1} \cdots f_{\theta_M}$}{
		      Identify $\gamma_m$, the set of indices of meta-validation batches, assigned to $f_{\theta_m}$;\\
		      Compute meta-test loss $\mathcal{G}_{\theta^{\prime}_m}$ using eq.~\eqref{eq:meta-test loss};\\
		      Perform meta-update using eq.~\eqref{eq:outer loop upate}
		      }
		  }
		 
	\caption{\muld~training}\label{algo:algo2}
\end{algorithm}

\subsection{Meta-test Stage}
\label{sec:meta-test}
While the train stage is similar to conventional ERM, our goal is to build ensembles that implicitly identify optimal domain groups for improved generalization. To achieve this goal, we systematically regulate the gradient flow from the meta-validation batches to each of the constituent models based on a model relevance score. We denote a generic, model relevance scoring function by $\mathrm{S}\colon\mathcal{V}_k\times f_{\theta_m}\rightarrow\mathbb{R}^+[0,1]$ which scores the model $f_{\theta_m}$ for a mini-batch $\mathcal{V}_k \subset \mathcal{D}_k^{v}$ from the meta-validation dataset. We denote by $\beta_{km}$ the resulting 
score \ie,
\begin{equation}
 \beta_{km}=\mathrm{S}(\mathcal{V}_k,f_{\theta_m})   
\end{equation}We now discuss the optimization process and subsequently describe our proposed gradient-matching based MRS below. 

\vspace{0.1in}
\noindent \textbf{Optimization:} Intuitively, when the relevance score $S$ is high, one expects that taking a gradient step for $\theta_m$ based on $\mathcal{B}$ is highly likely to improve the performance on $\mathcal{V}_k$. Hence, we compute this relevance score between every pair of meta-validation mini-batches $\{\mathcal{V}_{1}, \ldots, \mathcal{V}_{K}\}$ and the models $\{f_{\theta_1}, \ldots, f_{\theta_M}\}$ to obtain the matrix $\boldsymbol{\beta} \in \mathbb{R}^{K \times M}$. By identifying the model with the largest $\beta_{km}$ value, one can assign the most relevant model for each validation mini-batch $\mathcal{V}_k$. The final step is to compute the meta-gradients for $\theta_m, \forall m$ \wrt~to their ``relevant'' domains and perform a gradient-through-gradient update. We denote the indices of meta-validation batches that are assigned to $f_{\theta_m}$ by $\gamma_m = \{j \in (1, \cdots, K)\}$, such that for each $j$, model $f_{\theta_m}$ provides the largest MRS. The meta-validation loss $\mathcal{G}_{\theta^{\prime}_m}$ of $f_{\theta^{\prime}_m}$ using the relevant validation batches, $\gamma_m$, can be written as
\begin{equation}
\mathcal{G}_{\theta^{\prime}_m} =  \sum_{\forall(\x_i,y_i) \in \{\mathcal{V}_j\}, j \in \gamma_m} \ell(f_{\theta^{\prime}_m}(\x_i),y_i),
\label{eq:meta-test loss}
\end{equation} where the definition of $\theta'_m$ comes from eq.~\ref{eqn: inner_loop_update}. The final meta-update of $f_{\theta_m}$ using a gradient-through-gradient optimization can be written as follows:
\begin{equation}
\hat{\theta}_{m}=\theta_{m}- \lambda \frac{\partial( \mathcal{L}_{\theta_m}+\eta \mathcal{G}_{\theta^{\prime}_m})}{\partial \theta_{m}}.
\label{eq:outer loop upate}
\end{equation}
As a consequence of computing this assignment for meta-update in every iteration: (i) different parts of data from one source domain could get assigned to different models, thus producing a multi-domain ensemble that is guided by the inferred domain groups; (ii) in cases where none of the validation batches are assigned to a model, it converges to a standard ERM-based solution from eq.~\eqref{eqn: inner_loop_update}, which is still a strong baseline model. 

\vspace{0.1in}
\noindent \textbf{MRS Design:} As outlined above, through the MRS scoring function $\mathrm{S}$, we determine assignment of a model for a given meta-validation batch. We considered the following choices for implementing $\mathrm{S}$:
\begin{enumerate}
    \item \textit{Random Assignment:} In this case, $\mathrm{S}$ assigns randomly assigns each mini-batch to one of the models; 
    \item \textit{All-to-All assignment}: Here, every validation mini-batch $\mathcal{V}_k$ is assigned to all members of members;
    \item \textit{Loss-based assignment}: In this case, the empirical loss is used to determine the member assignment \ie, 
    $\beta_{km}=1-\frac{1}{|\mathcal{V}_k|} \sum_{(\x_i,y_i) \in \mathcal{V}_k} \ell(f_{\theta_m}(\x_{i}), y_{i})$.
    \item \textit{Gradient-matching}: While empirical loss-based assignment is a reasonable choice, we propose to implement MRS through gradient-matching. 
    \begin{definition}[Gradient-matching for model assignment]
\label{def:BMRS}

\begin{equation}
\label{eqn:BMRS}
\beta_{km}=\sum\nabla_{\theta_{m}}(\mathcal{L}_{\theta_m}).\nabla_{\theta_{m}}(\mathcal{G}_{\theta_m}^k),
\end{equation}
where $\mathcal{L}_{\theta_m}$ and $\mathcal{G}_{\theta_m}^k$ are the empirical risks (eq. ~\ref{eq:erm}) computed using the model $f_{\theta_m}$ on the meta-train ($\mathcal{B}$) and meta-validation ($\mathcal{V}_k$) batches respectively.
\end{definition}The summation is over all parameters in $\theta_m$, and this score computes the dot product between parameter sensitivities of $\theta_m$ \wrt~the train and validation batches. Though gradient-matching has been used in different contexts - for example, model alignment in MAML~\citep{finn2017model}, task affinity in multi-task learning~\citep{standley2020tasks}, promoting diversity in active sample selection~\citep{BADGE} etc., we make a surprising finding that gradient-based domain grouping is superior to loss-based grouping in domain re-labeling, though the latter metric is routinely used for model selection in MDG.
\end{enumerate}

\subsection{Model Selection Strategies}
A crucial component of any ZS-MDG algorithm is the specification of a model selection criterion. Here, model selection mainly refers to the selection of appropriate training checkpoints to evaluate on the unobserved domain $\mathcal{D^\dagger}$. It was found in~\citep{gulrajani2020search} that different model selection criteria lead to drastically different performance for the same method, thus making benchmarking of ZS-MDG approaches challenging.

In the context of MDG with ensemble-based approaches, the choice of model selection strategy has not been studied before. Note that, with \muld, we perform inference for a test sample by averaging the predictions from all $M$ models in the ensemble, $\x \in \mathcal{D^\dagger}$ , $\hat{y} = \frac{1}{ M}\sum_{m=1}^{M}f_{\theta_m}(\x)$. Defining the accuracy function as $\mathrm{A}: \x \times y \rightarrow \mathbb{R}^{+}[0,1]$, we can compute the performance of an individual model for a domain $k$ as $\mathrm{A}(\mathcal{D}_k^{hv}; f_{\theta_m})$ and that of an ensemble as $\mathrm{A}(\mathcal{D}_k^{hv}; \{f_{\theta_m}\})$. We investigate two model selection strategies in this study:

\noindent\textbf{Overall Avg:}
In this strategy, we choose the checkpoint in which each individual model $f_{\theta_m}$ produces high accuracy on each of the $K$ domains, on average. In other words,
$$
\arg \max_c \frac{1}{MK}\sum_{m=1}^M \sum_{k=1}^K  \mathrm{A}(\mathcal{D}_k^{hv}; f_{\theta_m}^c),
$$where $c$ indicates the training checkpoint index.

\noindent\textbf{Overall Ens:}
In this case, we choose the checkpoint in which the ensemble produces the highest accuracy for each of the $K$ domains, on average.
$$
\arg \max_c \frac{1}{K} \sum_{k=1}^K  \mathrm{A}(\mathcal{D}_k^{hv}; \{f_{\theta_m}^c\}).
$$

\subsection{Augmenting Meta-Validation Data} Our empirical study shows that by including synthetically augmented versions of the meta-validation batches, one can better leverage the intra-diversity in the source domains and further enhance the performance of \muld. Formally, we use standard image augmentation strategies (details in Section \ref{sec:experiments}) on the meta-validation batches $\{\mathcal{V}_1,\ldots,\mathcal{V}_K\}$ to produce  $\{\mathcal{V}_{K+1},\ldots,\mathcal{V}_{\bar{K}}\}$, where $\bar{K}-K$ is the number of additional batches. Note, through ablation studies, we demonstrate what role this augmentation for meta-validation data plays alongside the standard practice of training data augmentation adopted in the meta-train stage.

\section{Experiments}
\label{sec:experiments}
\begin{figure*}[t]
	\centering
	\includegraphics[width=0.95\linewidth]{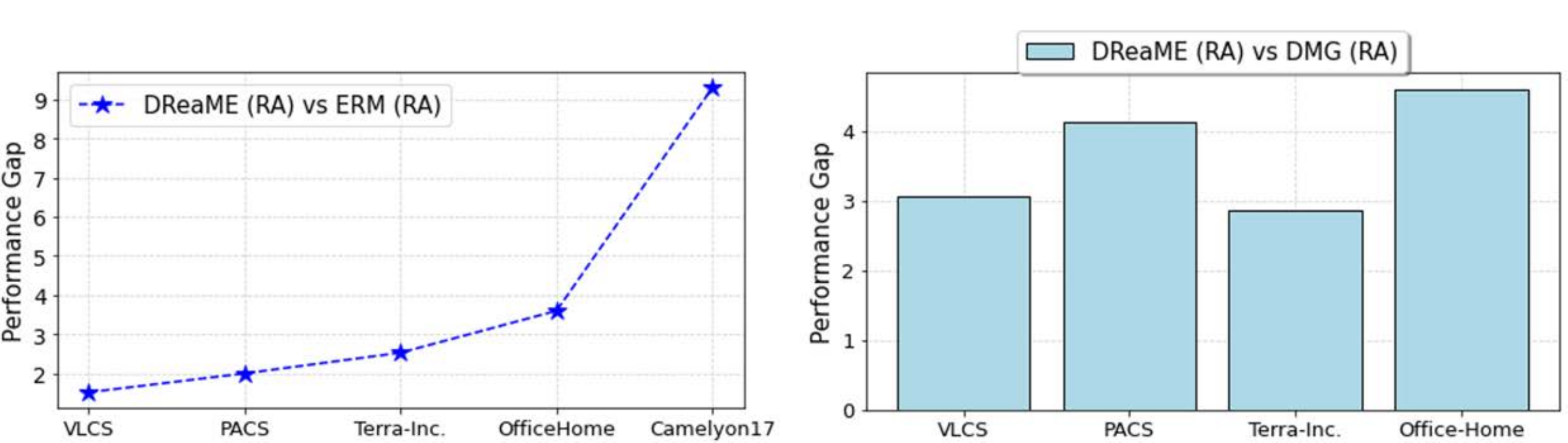}
	\caption{\textbf{Benchmarking \muld}. Our proposed approach significantly improves upon ERM (left) as well as sophisticated ensemble construction methods (right) in ZS-MDG, wherein we obtain larger performance gains as the domain discrepancy becomes more severe.}
	\label{fig:benchmark}
\end{figure*} 
\subsection{Dataset Description}
We evaluate \muld~using six standard visual MDG benchmarks (i) PACS~\citep{PACS} dataset comprising $4$ domains, namely photos, art, cartoon and sketches, with images belonging to $7$ different classes; (ii) VLCS~\citep{VLCS} dataset, which is also comprised of $4$ domains corresponding to the four benchmark image datasets (Caltech101, LabelMe, SUN09 and VOC2007) and contains images from $5$ classes; (iii) OfficeHome~\citep{officehome} dataset containing images from $65$ classes, where the images represent $4$ different domains, namely art, clipart, product and real respectively; (iv) Terra Incognita~\citep{terra} comprised of camera trap images of wild animals obtained from four different camera angles (\textit{i.e.}, domains) and $10$ different wildlife categories; and (v) Camelyon$17$-WILDS~\citep{bandi2018detection, koh2021wilds} consisting roughly 400k images of potentially cancer cells taken at different hospitals and scanners.  
\subsection{Experimental Setup}
Following standard practice in ZS-MDG, for every dataset except for Camelyon$17$-WILDS, we run experiments by leaving out one of $K$ domains for testing while using the $K-1$ domains for training.
For Camelyon$17$-WILDS~\citep{bandi2018detection, koh2021wilds}, we use the standard protocol of using data from first three hospitals as training domains and use data from fourth and fifth hospitals as validation and testing domains. To enable a fair comparison with the state-of-the-art, we use ResNet-50~\citep{he2016deep}, pre-trained on ImageNet~\citep{ILSVRC15} as the backbone feature extractor for all experiments.

For \regmethod, we use the following settings: (i) number of training iterations $n_{iter}$ is set to $5000$ and $T$ to $300$; (ii) number of groups is fixed at $K=4$; (iii) $\lambda,\gamma,\eta_q$ are set to $0.1$, $0.3$ and $0.2$ respectively; (iv) batch size of $32$ per domain; and (v) Adam optimizer~\citep{ADAM} with learning rate $5e-5$.

For \muld, we use a random 80-20 split from each of the source domains to obtain the train and validation sets, while the train set itself is further subdivided (80-20) to construct meta-train and meta-validation data. We report the mean and standard deviation of performance, obtained across three trials with different random seeds, for each experiment similar to~\citep{gulrajani2020search}. 
Across all experiments we use the following hyper-parameters: (i) batch size of $32$ per domain; (ii) Adam optimizer~\citep{ADAM} with learning rate of $5e-5$ (for both $\alpha$ and $\lambda$); (iii) number of training iterations set to $5000$ and (iv) ensemble size $M$ is set to $3$. We implement \muld~into the publicly available DomainBed framework.

\begin{figure*}[t]
	\centering
	\includegraphics[width=0.95\linewidth]{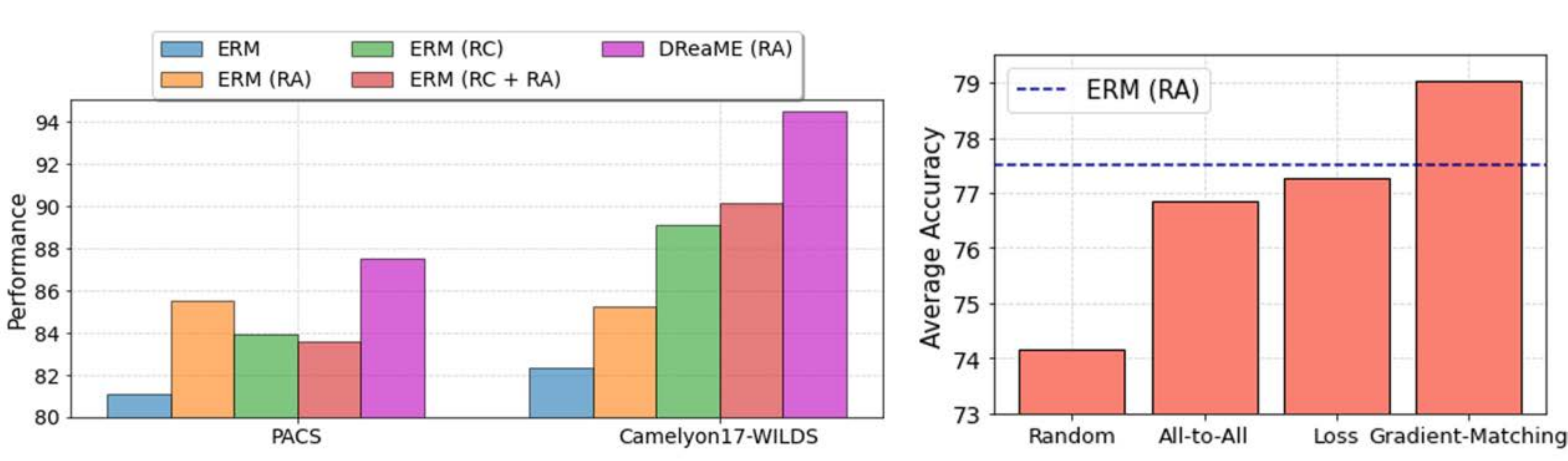}
	\caption{\textbf{Analysis of \muld~design}. (left) When compared to approaches that utilize advanced data augmentation strategies to improve generalization, \muld~eliminates the need for tailoring the augmentation strategy and is consistently effective for all benchmarks. In contrast, even sophisticated approaches such as RandConv (RC) or its combination with RandAug (RA)  provide varying degrees of improvements over ERM (RA) across different datasets; (b) When compared against different design choices for the MRS function $\mathrm{S}$, we find that the proposed gradient-matching performs the best.}
	\label{fig:behavior}
\end{figure*}
In \muld, the training mini-batches are augmented using a composition of the following augmentation choices: random horizontal flip, random color jitter and grayscaling with $10 \%$ probability, which we refer to as RandAug (RA). As described in Section~\ref{sec:meta-test}, we also create additional meta-validation batches by augmenting each batch $\mathcal{V}_k$ using subsets of augmentations used during training. We set $\eta$ in eq. \eqref{eq:outer loop upate} to $1.0$ and study the sensitivity of this hyper-parameter as part of our ablation study. We report results for both the proposed model selection strategies and our rigorous empirical study shows that the \textit{Overall Avg.} strategy provides a small margin of improvement over \textit{Overall Ens}. 
\begin{table*}[h]
\centering
\caption{\textbf{Summary performance of popular ZS-MDG baselines obtained using GroupDRO++ and \muld.} While the proposed GroupDRO++ improves over ERM and vanilla GroupDRO methods, \muld~with Overall (Avg.) model selection consistently achieves the best generalization performance.
}
\label{tab:datasets}
\renewcommand{\arraystretch}{1.2}
\resizebox{1.0\textwidth}{!}{ 
\begin{tabular}{|c||c||c||c||c|}

\rowcolor[HTML]{404040}

\textcolor{white}{\textbf{Methods}} & \textcolor{white}{\textbf{PACS}} & \textcolor{white}{\textbf{VLCS}} & \textcolor{white}{\textbf{OfficeHome}} & \textcolor{white}{\textbf{TerraIncognita}} \\ 
\hhline{=::=::=::=::=:}

ERM (RA)         & 85.5 $\pm$ 0.2     & 77.5 $\pm$ 0.4     & 66.5 $\pm$ 0.3            & 46.1 $\pm$ 1.8\\ \hhline{=::=::=::=::=:}
 
IRM~\citep{arjovsky2019invariant}                & 83.5 $\pm$ 0.8     & 78.5 $\pm$ 0.5     & 64.3 $\pm$ 2.2             & 47.6 $\pm$ 0.8 \\        \hhline{=::=::=::=::=:} 
MLDG~\citep{mldg}              & 84.9 $\pm$ 1.0     & 77.2 $\pm$ 0.4     & 66.8 $\pm$ 0.6             & \cellcolor[HTML]{FFDD86}47.7 $\pm$ 0.9            \\ \hhline{=::=::=::=::=:}
ARM~\citep{ARM}        & 85.1 $\pm$ 0.4      & 77.6 $\pm$ 0.3      & 64.8 $\pm$ 0.3             & 45.5 $\pm$ 0.3               \\ \hhline{=::=::=::=::=:}
RSC~\citep{RSC}              & 85.2 $\pm$ 0.9      & 77.1 $\pm$ 0.5      & 65.5 $\pm$ 0.9             & 46.6 $\pm$ 1.0          \\ \hhline{=::=::=::=::=:}
 
CORAL~\citep{CORAL}           & 86.2 $\pm$ 0.3     & 78.8 $\pm$ 0.6     & \cellcolor[HTML]{FFDD86}68.7 $\pm$ 0.3        & 47.6 $\pm$ 1.0            \\ \hhline{=::=::=::=::=:}

GroupDRO~\citep{GroupDRO} & 84.4 $\pm$ 0.8   & 76.7 $\pm$ 0.6      & 66.0 $\pm$ 0.7             & 43.2 $\pm$ 1.1              \\ 
\hhline{=::=::=::=::=:}
\hline\hline
\regmethod (ours) &  \cellcolor[HTML]{FFDD86}86.66 $\pm$ 0.4 & \cellcolor[HTML]{b1e9b0}\textbf{79.81 $\pm$ 0.5}& 67.1$\pm$0.3 & 47.45 $\pm$ 0.3 \\\hhline{=::=::=::=::=:}

 \muld~(Avg.) (ours) & \cellcolor[HTML]{b1e9b0} \textbf{87.35 $\pm$ 0.2} & \cellcolor[HTML]{FFDD86} 79.02 $\pm$ 0.3 & \cellcolor[HTML]{b1e9b0}\textbf{69.76 $\pm$ 0.2} & \cellcolor[HTML]{b1e9b0}\textbf{48.66 $\pm$ 0.2} \\
 \hline 
\end{tabular}
}
\end{table*}


\begin{table}[t]
\centering
\caption{\textbf{Performance of \muld~on the challenging WILDS-Camelyon17 benchmark}. In addition to being significantly superior to ERM (RA), \muld~outperforms best-performing methods such as SagNet~\citep{SagNet} and CORAL~\citep{CORAL}.}
\renewcommand{\arraystretch}{1.2}
\resizebox{0.5\textwidth}{!}{ 
\begin{tabular}{|c|c|}
\hline
\rowcolor[HTML]{404040}
\textcolor{white}{Method} & \textcolor{white}{Accuracy} \\ \hline \hline
ERM                     & 82.31          \\ 
ERM (RA)                     & 85.21                      \\
CORAL~\citep{CORAL} & 92.7\\
SagNet~\citep{SagNet} &  \cellcolor[HTML]{FFDD86}92.9\\
\hline \hline
GroupDRO++ (ours) & 86.7 \\
\muld~(Avg.) (ours)                   & \cellcolor[HTML]{b1e9b0} 94.6                           \\ \hline
\end{tabular}
}
\label{tab:wilds}
\end{table}

\subsection{Key Findings}
We now present a summary of the key findings from our empirical studies. We first find that re-labeling indeed leads to significant improvements over ERM thus providing evidence to our core hypothesis and sheds new light on the competitive behavior of ERM~\citep{gulrajani2020search}. From the results, we notice that \gro, despite leveraging the domain labels, often performs poorly compared to the vanilla ERM. In contrast, \regmethod improves upon both these baselines. Finally, we how that \muld~outperforms other existing ensembling strategies and produces state-of-the-art domain generalization performance.
\begin{table*}[h]
\centering
\caption{\textbf{Re-labeling domains improves generalization}. Here, for each dataset, we show the detailed generalization results for each of the domains using models trained with the remaining three domains. Both \muld~and \regmethod, which re-label samples from the different source domains and perform multi-domain training, lead to significant performance gains over ERM as well as the standard GroupDRO implementation.}
\label{tab:groupdro}
\renewcommand{\arraystretch}{1.2}
\resizebox{0.8\textwidth}{!}{ 
\begin{tabular}{|cccccc|}
\hline 
\rowcolor{gray!20}
\multicolumn{6}{|l|}{\textbf{Dataset: PACS}} \\
\hline 
\rowcolor[HTML]{404040}
\textcolor{white}{ Method} & \textcolor{white}{\textbf{A}} & \textcolor{white}{\textbf{C}} & \textcolor{white}{\textbf{P}} & \textcolor{white}{\textbf{S} }& \textcolor{white}{\textbf{Average}} \\
\hline 
ERM & $84.7 \pm 0.4$       & $80.8 \pm 0.6$  & $\cellcolor[HTML]{FFDD86}97.2 \pm 0.3$       & $79.3 \pm 1.0$       & $85.5$    \\ 
 \gro & $83.5 \pm 0.9$       & $79.1 \pm 0.6$  & $96.7 \pm 0.3$       & $78.3 \pm 2.0$       & $84.4$ \\
 \regmethod (ours) & \cellcolor[HTML]{FFDD86}${84.99\pm 0.2}$  & \cellcolor[HTML]{b1e9b0}$82.78\pm 0.4$  &  \cellcolor[HTML]{b1e9b0}$97.4\pm 0.3$ & \cellcolor[HTML]{FFDD86}${81.2\pm 0.6}$ & ${86.6 }$ \\
 \muld~(Avg.) (ours)& \cellcolor[HTML]{b1e9b0} $\cellcolor[HTML]{b1e9b0} 88.88\pm 1.0$  &\cellcolor[HTML]{FFDD86} $81.9\pm 1.7$  &  ${96.79\pm 0.4}$ & \cellcolor[HTML]{b1e9b0}$81.84\pm 0.1$ & \cellcolor[HTML]{b1e9b0}$87.35 $ \\
 
\hline \hline
\rowcolor{gray!20}
\multicolumn{6}{|l|}{\textbf{Dataset: VLCS}} \\
\hline 
\rowcolor[HTML]{404040}

\textcolor{white}{ Method} & \textcolor{white}{\textbf{C}} & \textcolor{white}{\textbf{L}} & \textcolor{white}{\textbf{S}} & \textcolor{white}{\textbf{V}} & \textcolor{white}{\textbf{Average}} \\
\hline 
ERM & $97.7\pm0.4$    &  $64.3 \pm 0.9$ & \cellcolor[HTML]{FFDD86}$73.4 \pm 0.5$  & $74.6 \pm 1.3$  &   $77.5 $     \\ 
 \gro & $97.3 \pm 0.3$ & $63.4 \pm 0.9$ & $69.5 \pm 0.8 $& $76.7 \pm 0.7$ & $76.7 $  \\
 \regmethod (ours) & \cellcolor[HTML]{FFDD86} $98.41 \pm 0.5$  &  $\cellcolor[HTML]{b1e9b0} 67.34 \pm 0.8$ &$\cellcolor[HTML]{b1e9b0}  75.7\pm 0.2$  & $\cellcolor[HTML]{FFDD86}{77.79 \pm 0.4}$& $\cellcolor[HTML]{b1e9b0} 79.81$   \\
 \muld~(Avg.) (ours)& $\cellcolor[HTML]{b1e9b0} 98.69\pm0.6$ &  \cellcolor[HTML]{FFDD86}$65.99\pm0.4$ & $73.12\pm0.5$& $\cellcolor[HTML]{b1e9b0} 78.27\pm0.6$ &\cellcolor[HTML]{FFDD86}$79.02$\\

\hline \hline
\rowcolor{gray!20}
\multicolumn{6}{|l|}{\textbf{Dataset: OfficeHome}} \\
\hline 
\rowcolor[HTML]{404040}
\textcolor{white}{Method} & \textcolor{white}{\textbf{A}} & \textcolor{white}{\textbf{C}} & \textcolor{white}{\textbf{P}} & \textcolor{white}{\textbf{R}} & \textcolor{white}{\textbf{Average}} \\
\hline 
ERM &  \cellcolor[HTML]{FFDD86} $61.30 \pm0.7$  & $52.40 \pm 0.3$  & $75.80 \pm 0.1$  & \cellcolor[HTML]{FFDD86}$76.60 \pm 0.3$  & $66.53$    \\ 
 \gro & $60.4 \pm 0.7$ & $52.7 \pm 1.0$ & $75.0 \pm 0.7$ & $76.0 \pm 0.7$  & $66.0$ \\
 \regmethod (ours) & $61.2 \pm 0.4$ & \cellcolor[HTML]{FFDD86} $54.4 \pm 0.2$ & \cellcolor[HTML]{FFDD86}$75.9 \pm 0.3$ & $75.02 \pm 0.2$ & \cellcolor[HTML]{FFDD86}$67.1$   \\
 \muld~(Avg.) (ours)& $\cellcolor[HTML]{b1e9b0} 64.38 \pm 1.2$ &$\cellcolor[HTML]{b1e9b0} 55.48 \pm 0.6$&$\cellcolor[HTML]{b1e9b0} 79.00 \pm 0.7$ & $\cellcolor[HTML]{b1e9b0} 80.16 \pm 0.4$&$\cellcolor[HTML]{b1e9b0} 69.76$ \\
\hline \hline
\rowcolor{gray!20}
\multicolumn{6}{|l|}{\textbf{Dataset: Terra Incognita}} \\
\hline 
\rowcolor[HTML]{404040}
 \textcolor{white}{Method} & \textcolor{white}{\textbf{L100}} & \textcolor{white}{\textbf{L38}} & \textcolor{white}{\textbf{L43}} &\textcolor{white}{ \textbf{L46}} & \textcolor{white}{\textbf{Average}}
 \\
\hline

 ERM & $49.8\pm4.4$ & $42.1\pm1.4$ & \cellcolor[HTML]{FFDD86}$56.9\pm1.8$ & $35.7\pm3.9$& $46.13$    \\ 
 \gro & $41.2\pm0.7$ & $38.6\pm2.1$ & $56.7\pm0.9$ & $36.4\pm2.1$& $43.2$ \\
 \regmethod (ours) & \cellcolor[HTML]{FFDD86}$50.7\pm0.9$ & $\cellcolor[HTML]{b1e9b0} 44.5\pm1.0$ & $\cellcolor[HTML]{b1e9b0} 57.4\pm 0.8$ & \cellcolor[HTML]{FFDD86} $37.2\pm 0.6$ & \cellcolor[HTML]{FFDD86}$47.45$\\
  \muld~(Avg.) (ours) & $\cellcolor[HTML]{b1e9b0} 53.65\pm2.9$& \cellcolor[HTML]{FFDD86}$44.25\pm1.7$ & $56.4\pm0.2$ & $\cellcolor[HTML]{b1e9b0} 40.35\pm0.6$ & $\cellcolor[HTML]{b1e9b0} 48.66$ \\
\hline 

 \end{tabular}}
\end{table*}

\vspace{0.05in}
\noindent\textbf{Finding 1: \regmethod and \muld~provide significant gains over ERM}.

\begin{figure*}[h]
    \centering
    \subfloat[]{
    \includegraphics[width=0.9\linewidth]{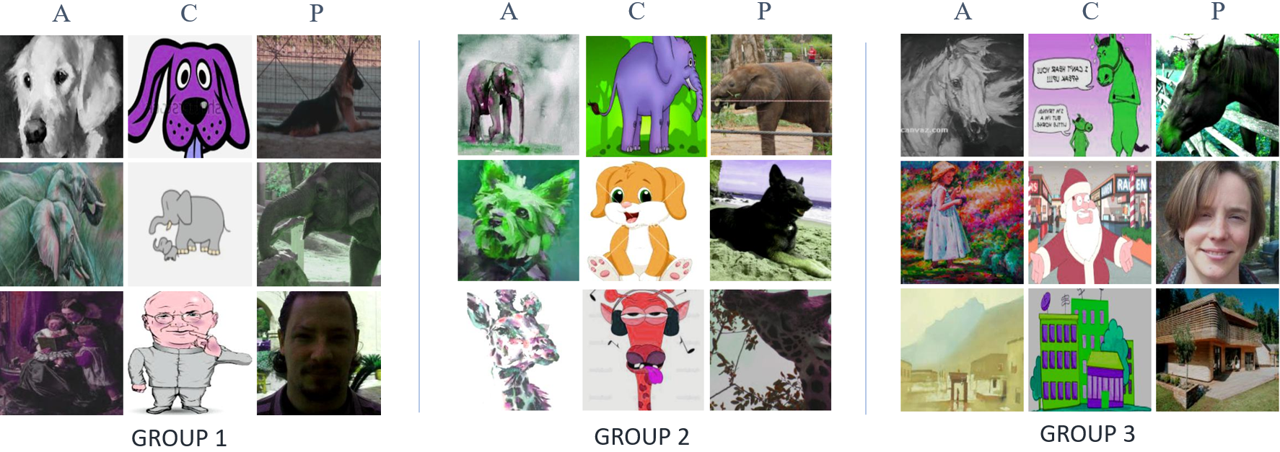}}
    
    \subfloat[]{
    \includegraphics[width=\linewidth]{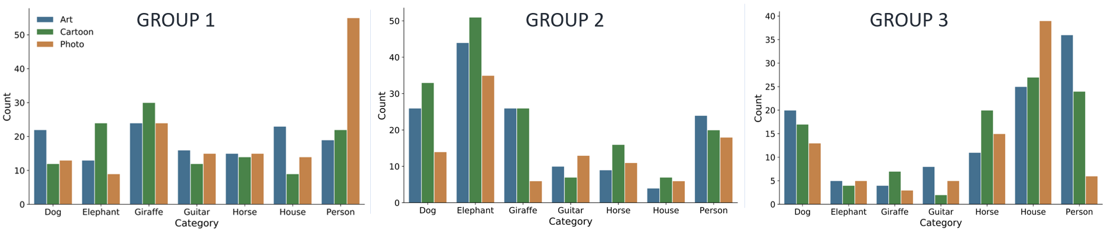}}
    
    \caption{\textbf{Visualizing the groups inferred using \muld}. (a) For this visualization, we trained \muld~using A (art), C (cartoon) and P (photo) domains from the PACS dataset and show randomly selected example images assigned to each of the three models in the ensemble. We notice that the MRS scoring based on gradient-matching assigns different subsets of each input domain to different groups; (b) We plot the distribution of class labels in each of the inferred domain groups. Interestingly, while MRS assigns different subsets of domains to different groups, it does not perform a trivial category split. In fact, our approach is able to effectively exploit the intra- and inter-domain discrepancies in order to evolve meaningful groups that maximally benefit the generalization performance.}
    \label{fig:groups}
\end{figure*}

\noindent As can be seen from Figure~\ref{fig:benchmark} (left), Tables~\ref{tab:datasets}, \ref{tab:groupdro} and \ref{tab:wilds}, using custom domain groups is highly beneficial. Overall, across the benchmarks, both GroupDRO++ and \muld~improve over ERM by significant margins, in terms of average generalization performance, regardless of the level of cross-domain gap inherent to each of the datasets. Furthermore, it is clearly apparent that GroupDRO++ is consistently better than the vanilla GroupDRO implementation. Through the use of a deep ensemble backbone, \muld~produces the best performance among the two proposed methods. For example, datasets such as TerraIncognita, and Camelyon$17$ are known to contain much higher cross-domain discrepancies compared to VLCS, thus making zero-shot generalization more challenging. However, we find that \muld~with $M = 3$ achieves large performance gains over ERM and importantly, the gap with ERM widens as the severity of domain shift increases. In the challenging Camelyon$17$-WILDS dataset, \muld~provides $>10\%$ gains over ERM. In order to gain insights into the behavior of \muld, in Figure \ref{fig:groups}, we visualize examples images assigned to the three models ($M = 3$) in the ensemble. For this illustration, we trained \muld~using the art (A), cartoon (C) and photo (P) domains from the PACS benchmark. This clearly shows that images from the same input domain can be assigned to different groups, thus enabling improved generalization. Interestingly, we observe that each model evolves to specialize for different styles (in terms of image statistics) and semantic concepts (e.g., images corresponding to the \textit{person} class are strongly representative of group 3).

\vspace{0.05in}
\noindent\textbf{Finding 2: \muld~outperforms existing ensembling methods with the same complexity}.

\noindent We perform a comparative analysis of \muld~to a state-of-the-art multi-domain ensembling method, DMG~\citep{DMG}, which infers individual masks over the neurons for each of the source domains. Note that both DMG and \muld~are implemented using the DomainBed framework, and hence use the same experiment protocol, \textit{i.e.}, architecture, augmentations etc.
In Figure~\ref{fig:benchmark}(right), the superiority of \muld~over DMG is clearly evident across all benchmarks, with improvements as high as $4.5\%$ on the Office-Home benchmark. Through our ERM style update in the meta-train stage and by taking into account the relevance of a member \wrt~to a domain during the meta-test stage, we effectively re-label the samples into optimal domain groups. Detailed results across datasets are reported in Table~\ref{tab:datasets}.

\vspace{0.05in}
\noindent\textbf{Finding 3: \muld~provides non-trivial improvements over SoTA DG methods}. 

\noindent Despite the effectiveness of ERM as a baseline, its inability to leverage domain discrepancies implies that there is a non-trivial performance gap between ERM and state-of-the-art MDG methods. Using rigorous comparisons with benchmarks created by~\citep{gulrajani2020search}, and compiled in Table~\ref{tab:datasets}, we find that \muld~is highly competitive with the SoTA methods, which rely on a variety of strategies to leverage cross-domain discrepancies. Interestingly, our approach produces state-of-the-art results on PACS (+$1.1\%$), VLCS(+$1.2\%$), Office-Home (+$1.36\%$) and matches the performance of the state-of-the-art SagNet on the TerraIncognita dataset. Similarly, as showed in Table~\ref{tab:wilds}, on Camelyon$17$, \muld~achieves an accuracy of $94.6\%$, outperforming best-performing approaches such as SagNet and CORAL by a margin of $\sim 2\%$.

\subsection{Ablations}
\label{sec:ablation}
In this section, we perform ablation experiments to understand (i) the role of our proposed gradient-matching based MRS; (ii) the impact of meta-validation and train augmentation; (iii) the choice of $\eta$ and lastly (iv) the impact of ensemble size on generalization performance.

\subsubsection{Choice of MRS strategy}As explained in Section 5, we experimented with four different choices for the scoring function $\mathrm{S}$. In Figure~\ref{fig:behavior}(right), we show the effect of these choices on the performance using the VLCS dataset. We find that the proposed gradient-matching based MRS performs the best, considerably outperforming loss-based model assignment.

\subsubsection{Impact of meta-validation and train augmentation} 

In Table~\ref{tab:aug}, we show how the performance of \muld~varies when the augmentation protocol is changed. As discussed earlier, we divide data from each observed domain into three disjoint sets - train, meta-validation and held-out validation. We explore whether adding synthetic augmentations to create additional meta-validation batches can lead to improved implicit grouping. Note that, in this setup, we applied standard augmentation (following DomainBed) to the meta-train batches. It is evident from the table that, removing meta-validation augmentation leads upto $~1.5 \%$ decrease in accuracy in the cases of VLCS and Office Home datasets.   

We performed another ablation by not including any augmentation to the meta-train batches, while still considering synthetically augmented meta-validation batches. We observed that, for benchmarks with larger cross-domain gap such as, TerraIncognita~\citep{terra}, removing augmentation during training has significant effect on performance with upto $~2.8 \%$ drop and unsurprisingly the performance in this case of no train augmentation is lower than ERM with train augmentation. In contrast, with benchmarks such as VLCS, not applying train augmentation does not show any apparent impact.
\begin{table*}[h!]
\caption{\textbf{Impact of augmenting meta-train and meta-validation sets}. Without meta-validation augmentation and only train augmentation, we observed reduced performance across all benchmarks. On the other hand, omitting train augmentation and including only meta-validation leads to bigger performance drops on more challenging datasets such as TerraIncognita. 
}
\label{tab:aug}
\centering
\adjustbox{max width=1.1\textwidth}{%
\renewcommand{\arraystretch}{1.2}
\begin{tabular}{|c|c|c|c|c|c|}
\hline
\rowcolor[HTML]{404040}
\multicolumn{2}{|c|}{\textcolor{white}{Method}}                                                                 & \textcolor{white}{PACS}       & \textcolor{white}{VLCS}       & \textcolor{white}{Office Home} & \textcolor{white}{Terra Incognita} \\ \hline \hline
\multirow{2}{*}{\begin{tabular}[c]{@{}c@{}}\muld~\\ (No Meta-Valid Aug)\end{tabular}} & Avg & 86.93 $\pm$ 0.73 & 78.50 $\pm$ 0.17 & 69.37 $\pm$ 0.12  & 48.38 $\pm$ 0.73      \\ 
                                                                                       & Ens & 86.58 $\pm$ 1.52 & 77.06 $\pm$ 0.05 & 68.74 $\pm$ 0.28  & 47.99 $\pm$ 0.97      \\ \hline
\multirow{2}{*}{\begin{tabular}[c]{@{}c@{}}\muld~\\ (No Train Aug)\end{tabular}}      & Avg & 85.46 $\pm$ 1.29 & 78.88 $\pm$ 0.45 & 69.05 $\pm$  0.28 & 45.87 $\pm$ 1.13      \\ 
                                                                                       & Ens & 85.21 $\pm$ 1.26 & 78.52 $\pm$ 0.27 & 69.29 $\pm$ 0.36  & 46.09 $\pm$ 0.52      \\ \hline
\multirow{2}{*}{\begin{tabular}[c]{@{}c@{}}\muld~\end{tabular}}      & Avg & 87.35 $\pm$ 0.22 & 79.02 $\pm$ 0.20 & 69.76 $\pm$  0.48 & 48.66 $\pm$ 0.85      \\ 
                                                                                       & Ens & 87.21 $\pm$ 0.94 & 78.40 $\pm$ 0.19 & 70.06 $\pm$ 0.16  & 48.48 $\pm$ 0.96      \\ \hline
\end{tabular}}
\end{table*}
\subsubsection{Choice of $\eta$}
Next, we studied the sensitivity of the penalty $\eta$, a hyper-parameter that controls the penalty for the meta-validation loss in eq.~\eqref{eq:outer loop upate} in terms of the performance of \muld~using the OfficeHome~\citep{officehome} dataset. As evidenced in Table~\ref{tab:eta}, for values greater than $0.2$ the performance of \muld~is stable with respect to changes in $\eta$.

\begin{table}[t]
    \centering
    \caption{\textbf{Impact of the choice of $\eta$}. We find that \muld~is stable \wrt~change in $\eta$ values.}
    \label{tab:eta}
    \renewcommand{\arraystretch}{1.2}
\begin{tabular}{|c|c|c|c|c|c|c|}
    \hline
    \rowcolor[HTML]{404040}
    \multicolumn{2}{|c|}{\textcolor{white}{Choice of $\eta$}}  & \textcolor{white}{A}          & \textcolor{white}{C}          & \textcolor{white}{P}          & \textcolor{white}{R}      & \textcolor{white}{Avg.}  \\
    \hline
    \hline
    \multirow{2}{*}{ $\eta=1$ } & Avg & 63.08 & 55.52 & 78.60 & 79.80 & 69.25 \\
    & Ens & 64.72 & 55.09 & 78.60 & 81.15 & 69.89 \\\hline
    \multirow{2}{*}{ $\eta=0.8$  } & Avg & 63.44 & 56.41 & 77.42 & 79.06 & 69.08 \\
    & Ens & 62.98 & 54.85 & 77.43 & 79.23 & 68.62 \\\hline
    \multirow{2}{*}{ $\eta=0.6$ } & Avg & 62.61 & 56.65 & 76.63 & 78.91 & 68.70 \\
    & Ens & 62.52 & 57.18 & 76.54 & 78.60 & 68.71 \\\hline
    \multirow{2}{*}{ $\eta=0.4$  } & Avg & 63.54 & 56.41 & 77.06 & 79.35 & 69.09 \\
    & Ens & 62.72 & 56.41 & 76.83 & 79.29 & 68.81 \\\hline
    \multirow{2}{*}{ $\eta=0.2$  } & Avg & 63.13 & 56.67 & 76.60 & 79.98 & 69.10 \\
    & Ens & 62.05 & 56.16 & 76.46 & 79.58 & 68.56 \\\hline
    \end{tabular}
\end{table}

\subsubsection{Impact of Ensemble Size.} The ensemble size $M$ is a standard hyper-parameter in any ensembling approach. The complexity (size) of the ensemble is controlled such that a simple averaging strategy can still work for unseen test domains. When $M$ becomes large and models become diverse, one might require a ``manager'' module to select a specific model from the ensemble. However, the ensemble size is not necessarily connected to the number of domains and for simplicity, we fixed $M=3$ for all cases (identified using parameter search on PACS, VLCS). Table~\ref{tab:change_m} shows the performance of \muld~on VLCS~\citep{VLCS} at different ensemble sizes ($M=2,3,4$).
We observe that while a three-model ensemble outperforms two-model ensemble by almost $2\%$, there is no significant improvement achieved beyond $M=3$. We note that at this time the choice of $M$ is empirical, and extending this framework to automatically identify the required number of models, such that simple averaging of predictions from the ensemble can still be effective, is part of our future work.
\begin{table}[h!]
    \centering
    \caption{\textbf{Impact of change in ensemble size}. For the VLCS benchmark, We observed that there was no improvement beyond $M=3$.}
    \label{tab:change_m}
    \renewcommand{\arraystretch}{1.2}
  
\begin{tabular}{|c|c|c|c|c|c|c|}
    \hline
     \rowcolor[HTML]{404040}
    \multicolumn{2}{|c|}{\textcolor{white}{Choice of M}}  & \textcolor{white}{C}          & \textcolor{white}{L}          & \textcolor{white}{S}          & \textcolor{white}{V}      & \textcolor{white}{Avg.}  \\
    \hline
    \hline
    \multirow{2}{*}{ M = 2 } & Avg & 98.32 & 62.58 & 69.04 & 77.78 & 76.93 \\
    & Ens & 98.05 & 62.40 & 72.88 & 76.78 & 77.53 \\\hline
    \multirow{2}{*}{ M = 3 } & Avg & 98.69  & 65.99  &73.12 &78.27 &79.02  \\
    & Ens & 98.60 & 65.61 &72.26 & 77.11 & 78.4\\\hline
    \multirow{2}{*}{ M = 4 } & Avg & 98.58 & 65.84 & 72.58 & 78.02 & 78.76 \\
    & Ens & 98.58 & 65.43 & 72.76 & 78.41 & 78.80 \\\hline
    \end{tabular}
\end{table}

\section{Related Work}
\label{sec:related_work}
In several computer vision tasks, deep models are often found to generalize poorly to out-of-distribution (OOD) data. Consequently, several problem formulations have emerged to address this critical limitation, for example, unsupervised domain adaptation, domain generalization and test-time adaptation. These approaches are typically differentiated by the assumptions made on the access to source (ID) data and target (OOD) data.  In this section, we provide a brief review of different multi-domain generalization approaches as it relates to our method. In particular, we organize existing methods broadly into methods that are domain-label agnostic and those that leverage the domain groups. 

\paragraph{Domain-Label Agnostic MDG}
This class of approaches ignores the domain labels and assumes the observed data from multiple sources to be drawn from a joint distribution. A prototypical example is the standard ERM~\citep{vapnik1999overview} training, which pools data from all source domains and infers a classifier that minimizes the risk on this pooled set. While this method has been long known to be inferior to domain-label aware methods, a recent large-scale study showed that with powerful feature extractors such as Resnet-50~\citep{he2016deep} and consistent model selection criteria, ERM is indeed a strong baseline. 
Huang \etal~\citep{RSC} propose to improve upon ERM by iteratively discarding the top $p$-percentile gradients in final layers of the network by zeroing them and then performing an update. They observed that by discarding top-$p$ gradients leads to better OOD generalization as compared to discarding top-$p$ features~\citep{park2016analysis}. Pezeshki \etal \citep{pezeshki2021gradient} further studied the interesting gradient starvation phenomenon that arises in ERM when optimized through gradient descent. They observed that ERM training captures only a subset of features while other predictive features are ignored, and hence spectral decoupling can be used to regularize the process.
\subsection{Domain-Label Aware MDG}
More recent approaches in MDG attempt to improve generalization by leveraging the prior knowledge about the domain groups (\eg photos, art, cartoon, sketch in PACS~\citep{PACS} dataset). At the core, these methods learn invariant features by decreasing the discrepancy between the different domains. For example, CORAL~\citep{CORAL} and CAADA~\citep{caada} simultaneously infer an effective classifier, while also matching the feature covariance matrices across different training domains either through customized loss functions or through adversarial learning. On the other hand, Gretton \etal~\citep{MMD} directly minimize the maximum mean discrepancy between different domains. 

In the pioneering work of DANN~\citep{DANN}, Ganin \etal utilized adversarial learning to learn domain-invariant features by constructing a gradient reversal layer. Several follow-up works~\citep{akuzawa2019adversarial, albuquerque2020adversarial} improved upon DANN by explicitly considering statistical dependence between domains. Style-Agnostic network~\citep{SagNet} is another recently proposed method that attempts to reduce the style bias corresponding to each domain by editing the styles through an adversarial learning paradigm. Building upon mixup~\citep{zhang2017mixup}, a popular approach in the image-classification setting, many methods~\citep{mixup_da,adversarial_mixup} have proposed inter-domain mixup \ie, learning from linearly interpolated examples from random pairs of domains and their labels. Marginal transfer learning (MTL)~\citep{blanchard2017domain,blanchardDG_thm} proposes to build classifiers that take in an additional input in the form of a domain-specific prototypical feature embedding. Adaptive risk minimization (ARM)~\citep{ARM} further improves upon MTL by using a separate embedding convolutional neural network. Arjovsky \etal~\citep{arjovsky2019invariant} subsequently introduced the paradigm of Invariant risk minimization (IRM) and proposed to learn feature representations such that there is a classifier that is simultaneously optimal for all domains. In general, causal learning has inspired MDG approaches such as IRM~\citep{arjovsky2019invariant}, MatchDG~\citep{mahajan2021domain} and Deep CAMA~\citep{zhang2020causal}.
 Meta-learning approaches such as MLDG ~\citep{mldg} build upon the MAML~\citep{MAML} framework to learn a feature extractor using only $K-1$ domains and the $K^{\text{th}}$ domain to perform the meta-update. 

Given the inherent limitations of ERM in handling multi-domain data~\citep{blanchardDG_thm}, researchers have explored the use of distributional robust optimization methods~\citep{GroupDRO,JTT,Vrex} that attempt to reduce the worst-group error \ie, the training domain with the highest error, by suitably weighting the loss function. In particular, \gro~\citep{GroupDRO} is a popular example that utilizes an adaptive sample weighting mechanism to suitably adjust the loss function. 

Finally, multi-domain ensembling forms an important class of MDG approaches that are often found to produce state-of-the-art generalization at the cost of the need to train and infer with multiple models. Existing multi-domain ensembling methods typically combine models independently trained on each of the source domains~\citep{dgens1,dg_ens2,DSON,DMG,matsuura2020domain}. For example,~\citep{DSON} uses a common model across the members of the ensemble, while allowing domain-specific normalization layers. On the other hand,~\citep{DMG} proposed to build domain-specific masks over neurons and optimized the masks for minimizing cross-domain feature overlap. Furthermore, other existing approaches have also focused on building an additional model to determine the weights to be used while aggregating the predictions from different models in an ensemble at test time~\citep{dofe}. While this approach could be seen as having a \textit{manager} from the Mixture of Experts literature, by design, we cannot employ such an approach when generalizing to an unseen domain in the zero-shot setting. This is due to the fact that in ZS-MDG problems, we can neither assume access to a collection of samples from the unseen domain, nor can we expect the \textit{manager} model to automatically withstand the complex distribution shifts. 





\section{Conclusion}
\label{sec:conclusions}
In this work, we explored the benefits of domain re-labeling in zero-shot multi-domain generalization. In particular, we considered two popular classes of MDG algorithms, namely distributional robust optimization and multi-domain ensembling, to demonstrate our hypothesis.
While we utilized an explicit clustering step to design \regmethod, through \muld~we proposed a meta-optimization strategy to produce multi-domain ensembles with implicit re-labeling. While the underlying MDG techniques and the strategies used for re-labeling are vastly different across these two solutions, we find that both the proposed approaches lead significant improvements in generalization performance over their standard implementations with original domain labels from the dataset. Using rigorous empirical studies on standard benchmarks, we find that both \regmethod and \muld~are able to outperform the ERM baseline, and on challenging datasets, achieve state-of-the-art MDG performance. Uncovering the theoretical underpinnings to understand the link between the choice of domain groups and MDG performance, and designing generic re-labeling strategies that can be readily integrated into any MDG method are crucial future research directions.  
\section*{Acknowledgements}
This work was performed under the auspices of the U.S. Department of Energy by the Lawrence Livermore National Laboratory under Contract No. DE-AC52-07NA27344, Lawrence Livermore National Security, LLC.

\vskip 0.2in
\bibliography{refs}

\end{document}